\documentclass{article}
\usepackage{spconf,amsmath,epsfig}
\usepackage{comment}
\usepackage{tabu}
\usepackage{subfig}

\newcommand{\method}{$OD2RNN$}

\title{Combining Sentinel-1 and Sentinel-2 time series via RNN for object-based Land Cover Classification}
\name{Dino Ienco,
      Raffaele Gaetano,
      Roberto Interdonato
      Kenji Ose
      and Dinh Ho Tong Minh
\sthanks{Thanks for the GEOSUD project with reference ANR-10-EQPX-20, the Programme National de T\'{e}l\'{e}d\'{e}tection Spatiale (PNTS, \url{http://www.insu.cnrs.fr/pnts} ), grant $n^{o}$PNTS-2018-5 for funding.}} 
\address{IRSTEA, UMR TETIS, LIRMM,  Univ. of Montpellier, Montpellier \\ 
CIRAD, UMR TETIS, Univ. of Montpellier, Montpellier \\
CIRAD, UMR TETIS, Univ. of Montpellier, Montpellier \\
IRSTEA, UMR TETIS, Univ. of Montpellier, Montpellier \\
IRSTEA, UMR TETIS, Univ. of Montpellier, Montpellier
}

\begin{document}
%
\maketitle
\begin{abstract}
Radar and Optical Satellite Image Time Series (SITS) are sources of information that are commonly employed to monitor earth surfaces for tasks related to ecology, agriculture, mobility, land management planning and land cover monitoring. Many studies have been conducted using one of the two sources, but how to smartly combine the complementary information provided by radar and optical SITS is still an open challenge. In this context, we propose a new neural architecture for the combination of Sentinel-1 (S1) and Sentinel-2 (S2) imagery at object level, applied to a real-world land cover classification task. 
Experiments carried out on the \textit{Reunion Island}, a overseas department of France
in the Indian Ocean, demonstrate the significance of our proposal.


\end{abstract}

\section{Introduction}
\label{sec:intro}
Modern Earth Observation (EO) systems provide huge volumes of  data every day, involving programs that produce freely available multi-sensor satellite images at high spatial resolution with low temporal revisit period. Satellite Image Time Series (SITS) are particularly useful for tasks such as land cover classification~\cite{IngladaVATMR17} and natural habitat monitoring~\cite{KoleckaGPPV18}. A notable example is represented by the two main ESA Sentinel missions (Sentinel-1 and -2) that provide optical (multi-spectral) and radar imagery at 10~m spatial resolution with revisit time of 6 and 5 days respectively. An open challenge in the remote sensing community~\cite{Benedetti18} is how to efficiently combine the complementary information coming from these sensors, namely, the properties of surface materials provided by the optical sensor (S2) and the structural characteristics of landscape elements provided by the radar sensor (S1).
In this context, we propose a new deep learning architecture, named \method{} (Object-Based two-Branch RNN Land Cover Classification), to manage multi-temporal and multi-sensor data (radar and optical satellite image time series) at object-level to leverage the complementarity between these two different types of information. Working at object level instead of pixels has two main advantages: i) objects represent more representative and potentially feature-rich pieces of information and ii) object based approaches facilitate data analysis scale-up since, for the same area, the number of objects is usually smaller than the number of pixels by several order of magnitude.

\section{Data Description}
\label{sec:data}
The analysis is carried out on the \textit{Reunion Island}, a French overseas department located in the Indian Ocean. The dataset consists of a time series of 34 S2 images and a time series of 24 S1 images both acquired between April 2016 and May 2017. 
For S2, we used level-2A products by the THEIA pole~\footnote{Data are available via \url{http://theia.cnes.fr}, preprocessed in surface reflectance via the \textit{MACCS-ATCOR Joint Algorithm}~\cite{Hagolle2015} developed by the National Centre for Space Studies (CNES).}. Here, we only use bands at 10m, in the blue, green, red and near infrared spectrum (resp. B2, B3, B4 and B8). A preprocessing was performed to fill cloudy observations through a linear multi-temporal interpolation over each band (cf.~\textit{Temporal Gapfilling}~\cite{IngladaVATMR17}), and the NDVI radiometric index was calculated for each date~\cite{IngladaVATMR17} (5 variables for each timestamp).
For S1, images are acquired in TOPS mode with dual-polarization (VV+VH). The backscatter images were generated and radiometrically calibrated 
using parameters included in the S1 SAR header, then coregistered with the S2 time series. The pixel size of the orthorectified image data is 10 m. After geocoding, all backscatter images are converted to the logarithm dB scale, normalized to values between 0-255 (8 bits).
The spatial extent of the \textit{Reunion Island} site is 6\,656 $\times$ 5\,913 pixels. The field database was built from various sources (see \cite{Benedetti18} for details) 
and is available in GIS vector format as a collection of class attributed polygons 
that have been, successively, converted in raster format at the S2 spatial resolution (10m). The final ground truth includes 2\,656 objects distributed over 13 classes (Table~\ref{tab:data_reu}).

\begin{table}[!htbp]
\centering
\footnotesize
\begin{tabular}{|l||c|c|c|}
	\hline
\textbf{Class} & Label & \# \textbf{Polygons} & \# \textbf{Objects} \\ 
\hline \hline
0 & {\em Crop cultivation} & 380 & 635 \\ \hline
1 & {\em Sugar cane} & 496 & 1205 \\ \hline
2 & {\em Orchards} & 299 & 640 \\ \hline
3 & {\em Forest plantations} & 67 & 167 \\ \hline
4 & {\em Meadow} & 257 & 786 \\ \hline
5 & {\em Forest} & 292 & 1016 \\ \hline
6 & {\em Shrubby savannah} & 371 & 739 \\ \hline
7 & {\em Herbaceous savannah} & 78 & 143 \\ \hline
8 & {\em Bare rocks} & 107 & 250 \\ \hline
9 & {\em Urbanized areas} & 125 & 1499 \\ \hline
10 & {\em Greenhouse crops}& 50 & 163 \\ \hline
11 & {\em Water surfaces} & 96 & 159 \\ \hline
12 & {\em Shadows} & 38 & 60 \\ \hline
\end{tabular}
\caption{Characteristics of the \textit{Reunion Island} Dataset\label{tab:data_reu}}
\end{table}

Our purpose is to perform an object-oriented analysis of the \textit{Reunion Island} exploiting the available ground truth data. To this purpose, we use a SPOT6/7 image, acquired on April 6th 2016 and originally consisting of a 1.5~m panchromatic band and 4 multispectral bands (blue, green, red and near infrared) at 6~m resolution. The multispectral image has been resampled at 10~m (the same resolution of S2 images) using bicubic interpolation. Finally, the resulting image has been segmented via the Large Scale Generic Region Merging~\cite{Lassalle15} remote module of the Orfeo Toolbox toolkit~\footnote{\url{https://www.orfeo-toolbox.org/CookBook/}} obtaining 167\,319 segments. The obtained segments where spatially intersected with the ground truth data to provide radiometrically homogeneous class samples and 7\,462 labeled segments of comparable sizes are finally obtained. 
To integrate information from the time series, each object is then attributed with the mean of the corresponding pixels over the selected bands and indices, (Blue, Green, Red, NIR and NDVI for S2, VV and VH for S1) for all the available timestamp, achieving a total of $34*5 + 24*2 = 218$ variables per object. 




\section{Object-Based two-Branch RNN Land Cover Classification}
\label{sec:method}
\begin{figure}[t]
\centering
\includegraphics[width=\columnwidth]{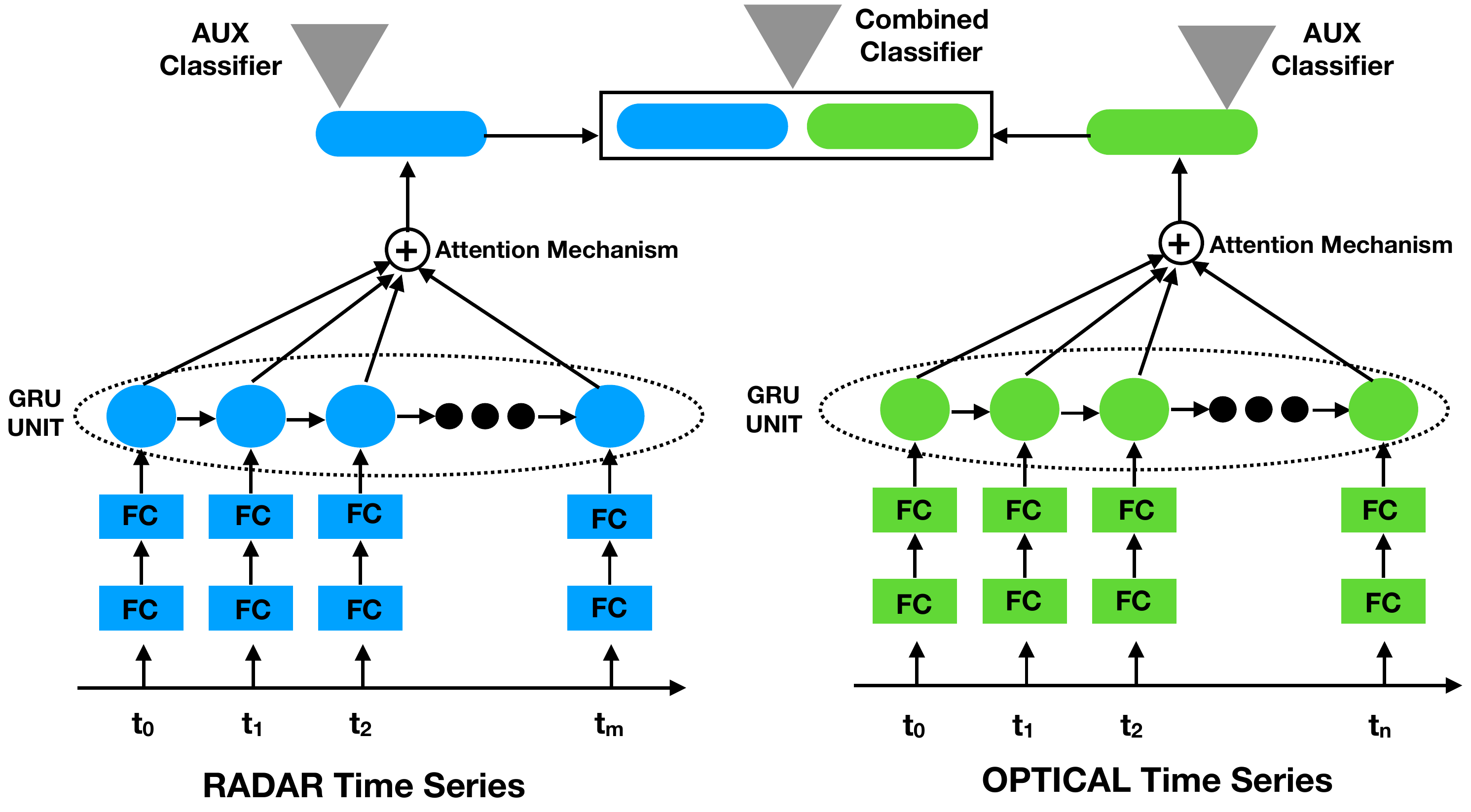}
\caption{ \label{fig:OD2RNN} Visual representation of the  \method{} Deep Learning Architecture.}
\end{figure}

Figure~\ref{fig:OD2RNN} depicts the proposed \method{} deep learning architecture for the multi-source satellite image time series classification process. 
Inspecting the model, We can observe that, structurally, our architecture has two twin streams: one for the radar time series and one for the optical time series. The output of the model is a land cover classification class for each pair (radar/optical) time series.
Each stream of the \method{} architecture can be roughly decomposed in three parts: i) data preprocessing and enrichment ii) time series analysis and iii) multi-temporal combination to generate per-source features. Finally, the radar (blue stream) and optical (green stream) information are combined together. More in detail, the feature extracted from each stream (rectangle with rounded corners) are concatenated and directly leveraged to produce the final land cover classification. We name such learned features $opt_{feat}$ (resp. $radar_{feat}$) to indicate the output of the optical (resp. radar) stream.

Considering each stream of our model, we can identify three main stages. The first one is represented by two stages of fully connected (FC) layers that take as input one time stamp of the object time series (radar or optical) and combine the input data. Such stage allows the architecture to extract an useful input combination for the classification task enriching the original data representation. The non-linear transformation associated to the FC layer is the ReLU non linear activation function~\cite{NairH10}.  The second stage is constituted by a Gated Recurrent Unit~\cite{ChoMGBBSB14} (GRU) a kind of Recurrent Neural Network. Unlike standard feed forward networks (i.e. CNNs), RNNs explicitly manage temporal data dependencies since the output of the neuron at time $t-1$ is used, together with the next input, to feed the neuron itself at time $t$. Furthermore, this approach explicitly models the temporal correlation present in the object time series and it is able to focus its analysis on the useful portion of the time series discarding useless information. The third and last stage is implemented via neural attention~\cite{BritzGL17} on top of the outputs produced by the GRU model. Attention mechanisms are widely used in automatic signal processing (1D signal or language) and they allow to join together the information extracted by the RNN model at different time stamps via a convex combination of the input sources. We adopt the same attention mechanism employed in~\cite{Benedetti18}. Finally, once each stream has processed the corresponding time series information, the concatenation of the extracted radar and optical features is used to perform the classification.
To further strengthen the complementarity as well as the discriminative power of the learned features for each stream, we adapt the technique proposed in~\cite{HouLW17}, including an auxiliary classifier for each set of learned features ($opt_{feat}$ and $radar_{feat}$). The aim is to stress the fact that the learned features need to be discriminative alone (i.e., independently from each other).. 
Then, the learning process involves the optimization of three classifiers at the same time: one specific to $opt_{feat}$, one related to $radar_{feat}$ and one that considers the concatenation $[radar_{feat},opt_{feat}]$. The cost function associated to our model is :

\begin{align}
L_{total} &= .5 * L_1(radar_{feat}) + .5 * L_2(opt_{feat}) \nonumber \\
          &+ L_{fus}([radar_{feat}, opt_{feat}]) \label{eqn:cost}
\end{align}

\noindent where $L_i(feat)$ is the loss function (in our case the Cross-Entropy function) associated to the classifier fed with the features $feat$. We empirically weight the contribution of the auxiliary classifier with a weight of .5 to enforce the discriminative power of the per-source learned features. The final land cover class is derived combining the three (Softmax) classifiers with the same weight schema employed in the learning process. In addition, dropout (with a drop rate equal to 0.4) is employed for the GRU unit and between the two Fully Connected layers. The model is learned end-to-end.

\section{Experimental Evaluation}
\label{sec:expe}
In this section, we present and discuss the experimental results obtained on the study site introduced in Section~\ref{sec:data}. To evaluate the behavior of \method{}, we compare its performance with the one of a Random Forest classifier learned over radar, optical and optical/radar data. We name such competitors: $RF(S1)$, $RF(S2)$ and $RF(S1,S2)$, respectively. The values are normalized, per band (resp. indices) considering the time series, in the interval $[0,1]$.

To learn \method{} parameters we use the Adam optimizer~\cite{KingmaB14} with a learning rate equal to $1 \times 10^{-4}$. The training process is conducted over 1000 epochs with a batch size equals to 32. On average, each train epoch takes 11 seconds. The number of hidden units for the RNN module is fixed to 1\,024 (resp. 512) for the optical (resp. radar) branch while, we employ 32 and 64 neurons for the first and the second Fully Connected layers for each stream. We divide the dataset into three parts: training, validation and  test set with a proportion of 50\%,20\% and 30\% of the objects, respectively. Training data are used to learn the model. The model that achieves the best accuracy on the validation set is subsequently employed to classify the test set. For the $RF$ models, we optimize the model via the maximum depth of each tree (in the range \{20,40,60,80,100\}) and the number of trees in the forest (in the set \{100, 200, 300,400,500\}). Experiments are carried out on a workstation with an Intel (R) Xeon (R) CPU E5-2667 v4@3.20Ghz with 256 GB of RAM and four TITAN X GPU. 
The assessment of the classification performances is done considering global precision (\textit{Accuracy}), \textit{F-Measure} and \textit{Kappa} measures. For each evaluation metric, we report results averaged over ten random splits performed with the previously presented strategy.

\begin{table}[!htbp]
\centering
\scriptsize
 \begin{tabular}{| l || c | c | c |} \hline
& \textit{ F-Measure} & \textit{ Kappa} &\textit{ Accuracy} \\ \hline
$RF(S1)$ & 65.80 $\pm$ 0.85 & 0.6297 $\pm$ 0.0096 & 68.03 $\pm$ 0.82 \\ \hline
$RF(S2)$ & 86.10 $\pm$ 0.72 & 0.8442 $\pm$ 0.0082 & 86.40 $\pm$ 0.71 \\ \hline
$RF(S1,S2)$ & 81.26 $\pm$ 0.85 & 0.7936 $\pm$ 0.0103 & 82.01 $\pm$ 0.90 \\ \hline
\method & \textbf{89.48} $\pm$ 0.36 & \textbf{0.8811} $\pm$ 0.0041 & \textbf{89.59} $\pm$ 0.36 \\ \hline
 \end{tabular}
 \caption{F-Measure, Kappa and Accuracy considering \method{} and different competing methods. \label{tab:ReunionGen}}
\end{table}

Table~\ref{tab:ReunionGen} reports the average results of the different methods on the \textit{Reunion Island} dataset. We can observe that, considering the average behavior, \method{} clearly outperforms all Random Forest approaches. The best competitor is represented by the model leveraging only the optical data ($RF(S2)$). Not surprisingly, using only radar information ($RF(S1)$) provides the worst performances considering the land cover nomenclature on the study site.
Interestingly, RF performance on the combined data ($RF(S1,S2)$) is lower than the one obtained exploiting only optical information ($RF(S2)$).

\begin{table*}[!htbp]
\centering
\begin{tabular}{|l||c|c|c|c|c|c|c|c|c|c|c|c|c|}
	\hline
\textbf{Method} & 0 & 1 & 2 & 3 & 4 & 5 & 6 & 7 & 8 & 9 & 10 & 11 & 12\\  \hline \hline
$RF(S1)$ & 56.15 & 80.04 & 58.16 & 40.76 & 78.15 & 63.27 & 46.95 & 45.51 & 78.85 & 75.8 & 12.25 & 81.44 & 0.0 \\ \hline
$RF(S2)$ & 74.79 & 90.95 & 81.29 & 77.15 & 81.92 & 88.67 & 85.78 & \textbf{74.16} & 82.34 & 94.19 & 60.29 & \textbf{93.94} & \textbf{93.02} \\ \hline
$RF(S1,S2)$ & 69.35 & 89.85 & 75.76 & 57.74 & 82.83 & 82.26 & 78.13 & 62.87 & 72.73 & 92.14 & 32.22 & 89.38 & 83.58 \\ \hline
\method & \textbf{84.27} & \textbf{93.64} & \textbf{87.69} & \textbf{82.61} & \textbf{90.8} & \textbf{89.82} & \textbf{87.27} & 73.72 & \textbf{83.93} & \textbf{94.94} & \textbf{68.44} & 91.37 & 82.31 \\ \hline
\end{tabular}
\caption{Per-Class \textit{F-Measure} \label{tab:PerClass_fm_reunion}}
\end{table*}

Table~\ref{tab:PerClass_fm_reunion} reports the per-class F-Measure obtained by each method. Here, we can observe that \method{} achieves the best performance on ten classes over thirteen. Of particular interest is the behavior obtained on the class $0$--\textit{Crop Cultivation}, where our approach shows a gain of almost $10$ points over the best competitor ($RF(S2)$).
Considering classes $7$--\textit{Herbaceous savannah}, $11$--\textit{Water surfaces} and $12$--\textit{Shadows}, the best scores are reported by $RF(S2)$. Regarding the first two classes, \method{} achieves comparable results w.r.t. $RF(S2)$, while on the latter the difference is higher. However, the $12$ --\textit{Shadows} class is not a real land cover class and it was manually introduced by field experts since some areas of the island are constantly covered by shadows at S2 acquisition times. Indeed, the nomenclature was set up for optical-based classification since the optical signal can be affected by shadowing effects while this is not the case for the radar. This also explains why $RF(S1)$ has serious issues on such particular class.

\begin{figure*}[!ht]
\centering
\begin{tabular}{cccc}
\textit{VHSR Image} & \textit{RF(S2)} & \textit{\textit{RF(S1S2)}} & \method{} \\

\subfloat[\label{fig:reunion_ex1_b}]{\includegraphics[width=.2\linewidth]{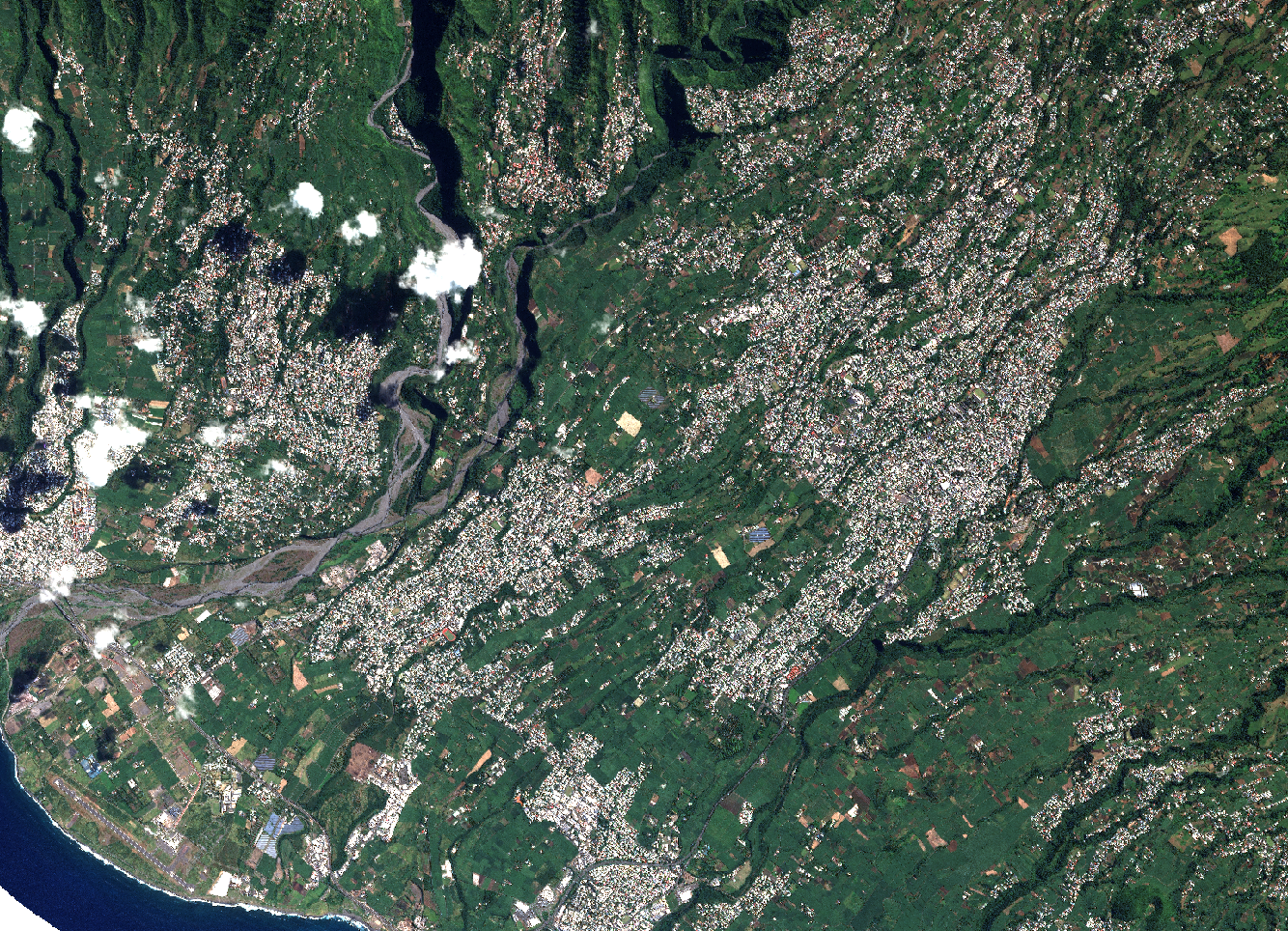} } & \subfloat[\label{fig:reunion_ex1_b}]{\includegraphics[width=.2\linewidth]{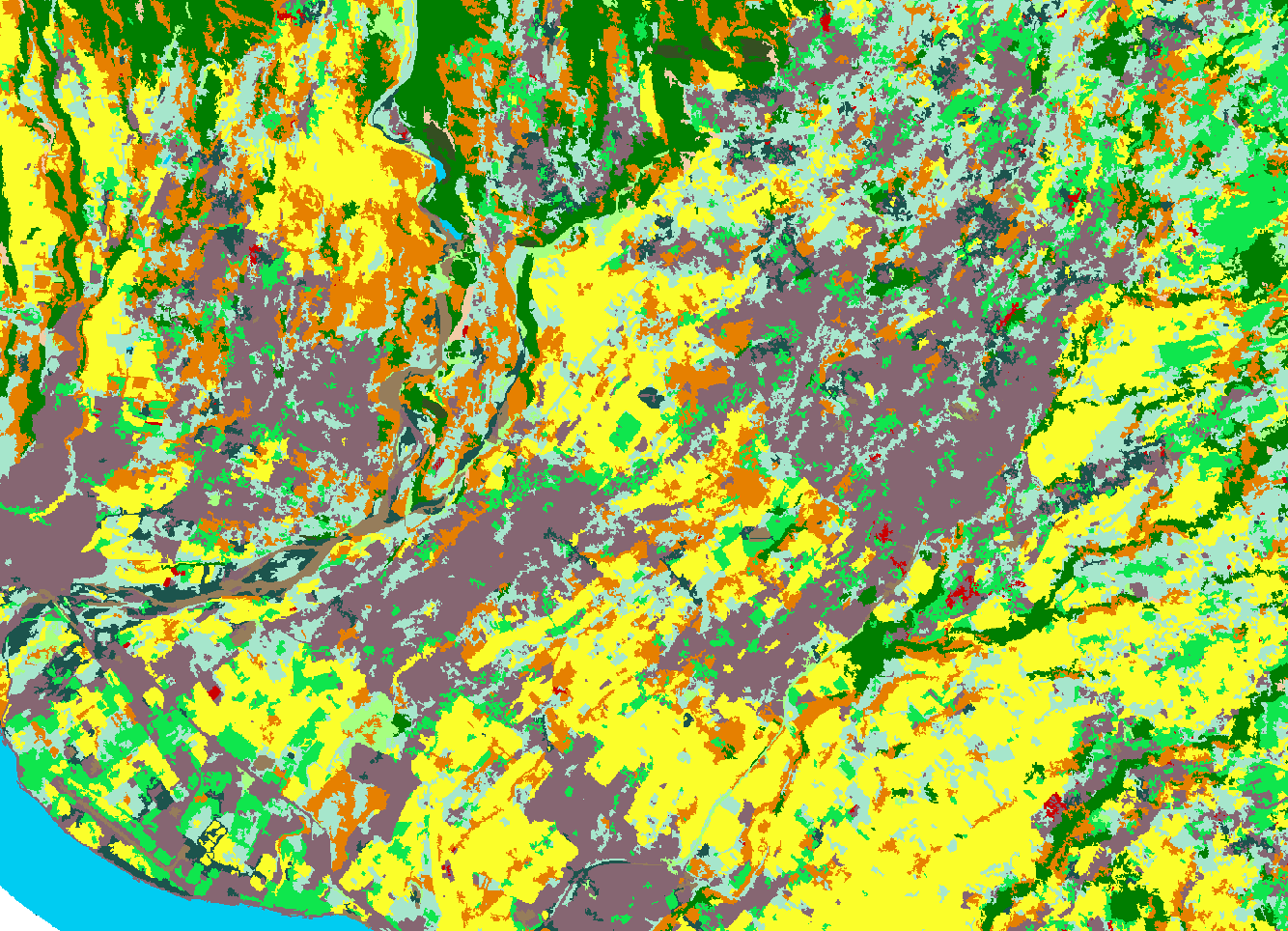} } & \subfloat[\label{fig:reunion_ex1_b}]{\includegraphics[width=.2\linewidth]{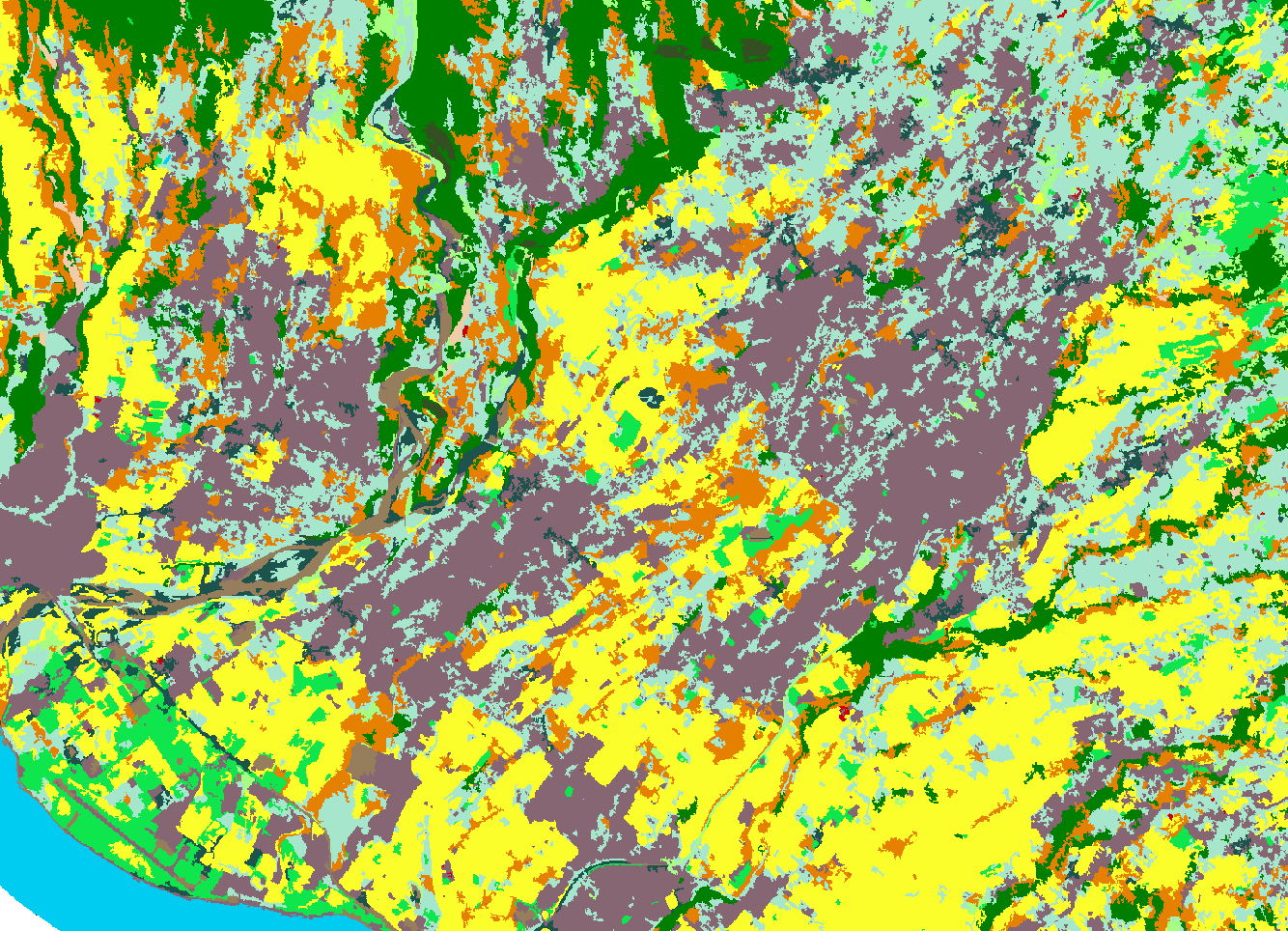} } & \subfloat[\label{fig:reunion_ex1_b}]{\includegraphics[width=.2\linewidth]{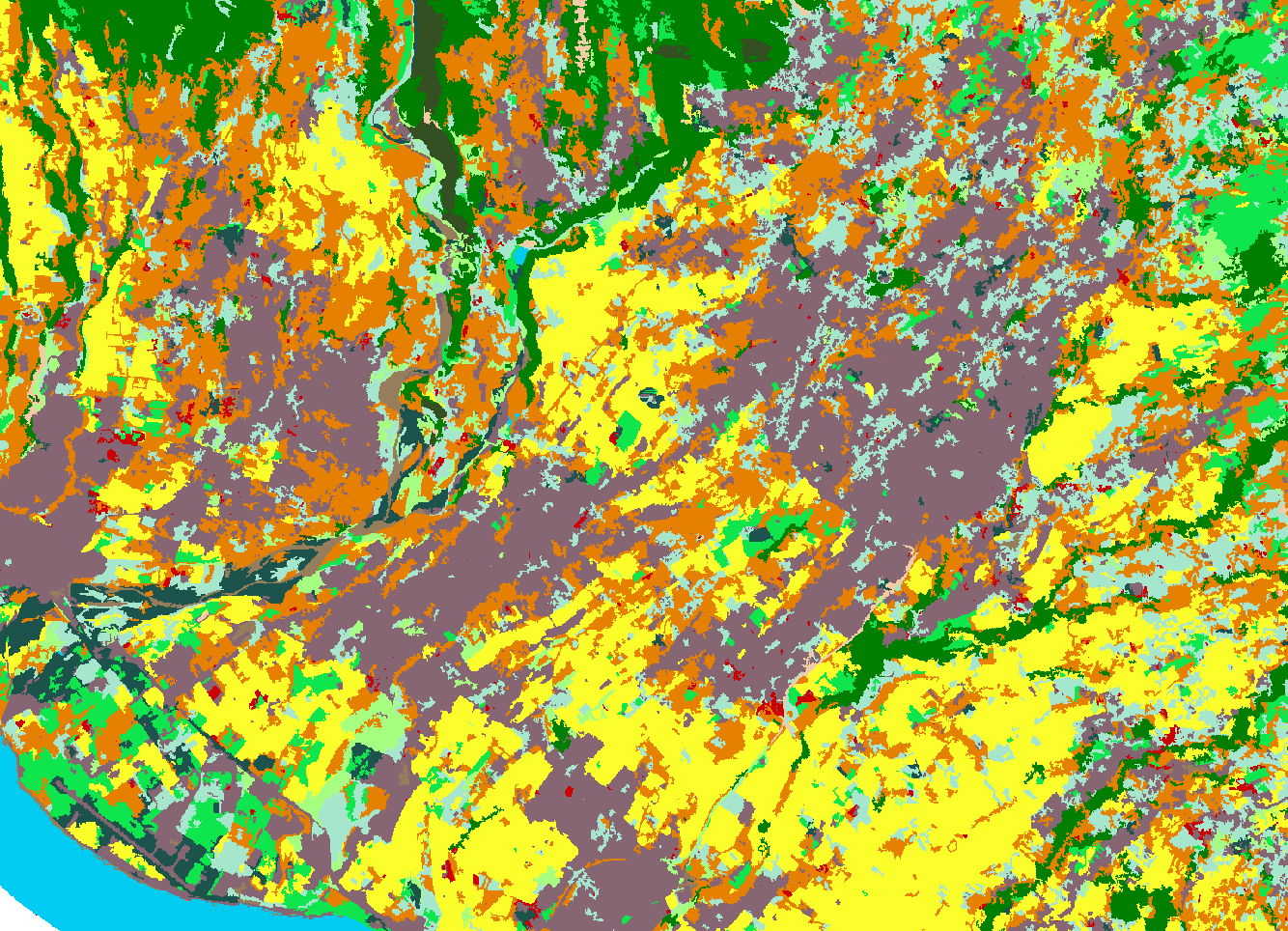} } \\

\subfloat[\label{fig:reunion_ex1_b}]{\includegraphics[width=.2\linewidth]{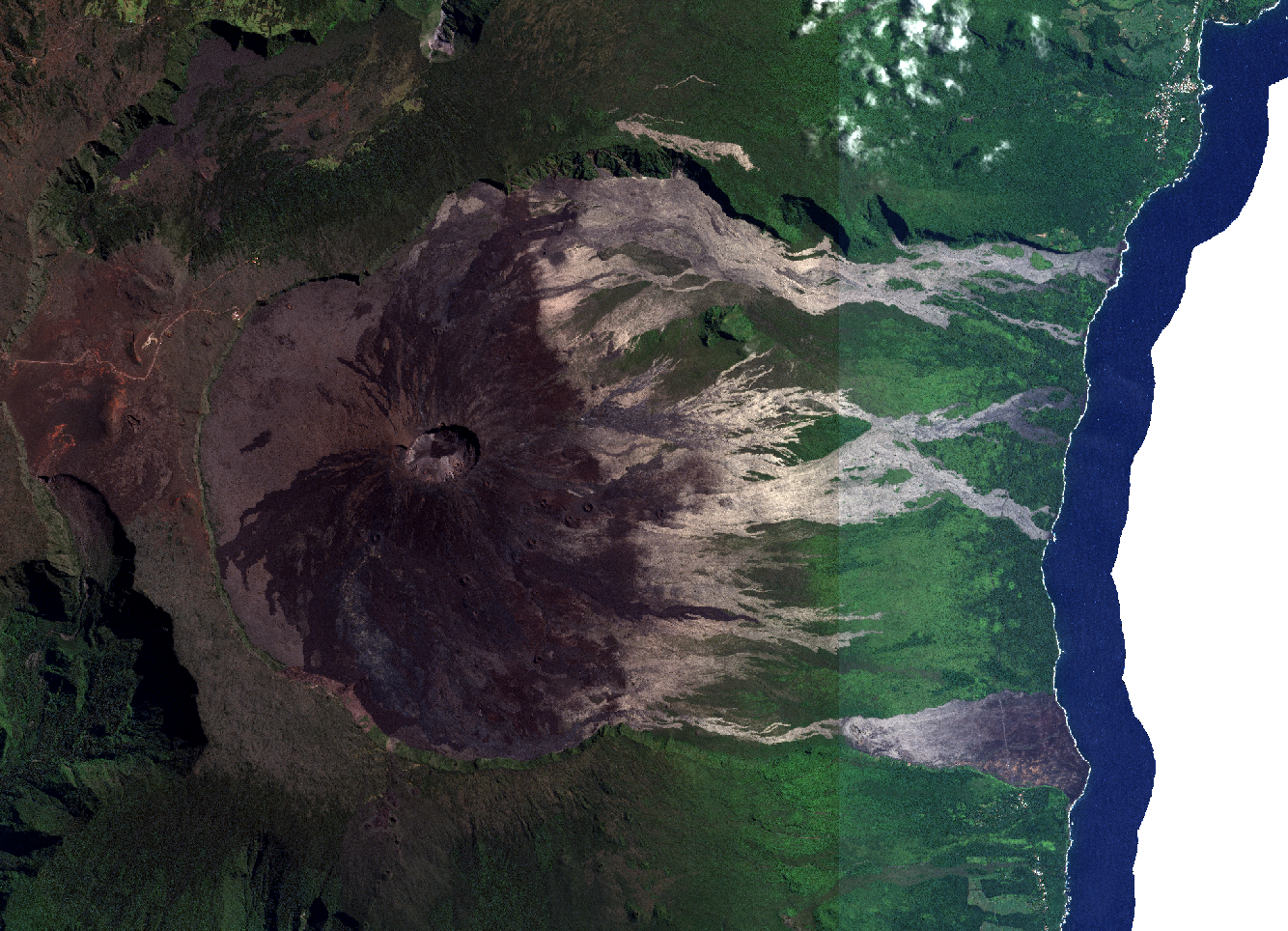} } & \subfloat[\label{fig:reunion_ex1_b}]{\includegraphics[width=.2\linewidth]{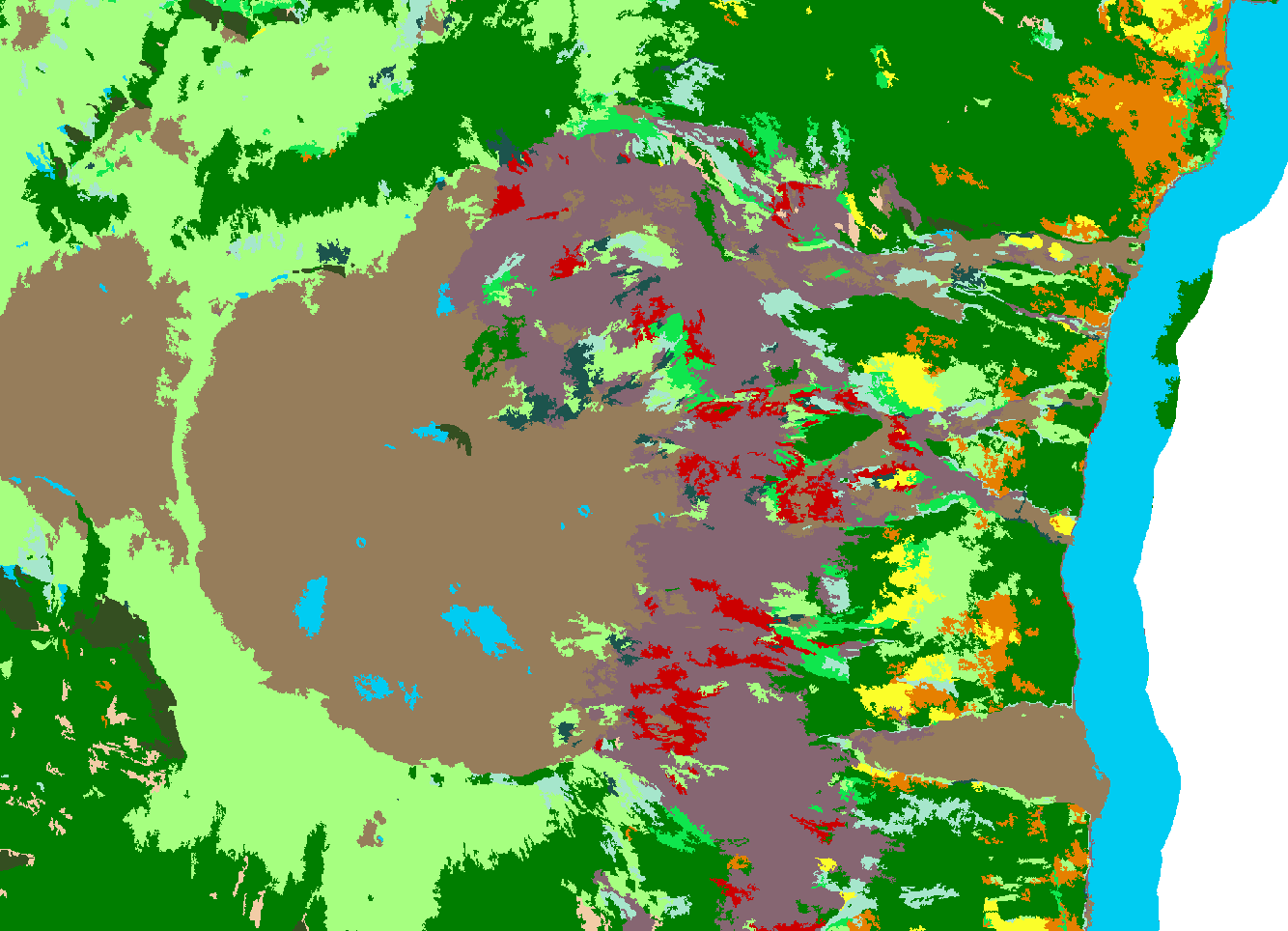} } & \subfloat[\label{fig:reunion_ex1_b}]{\includegraphics[width=.2\linewidth]{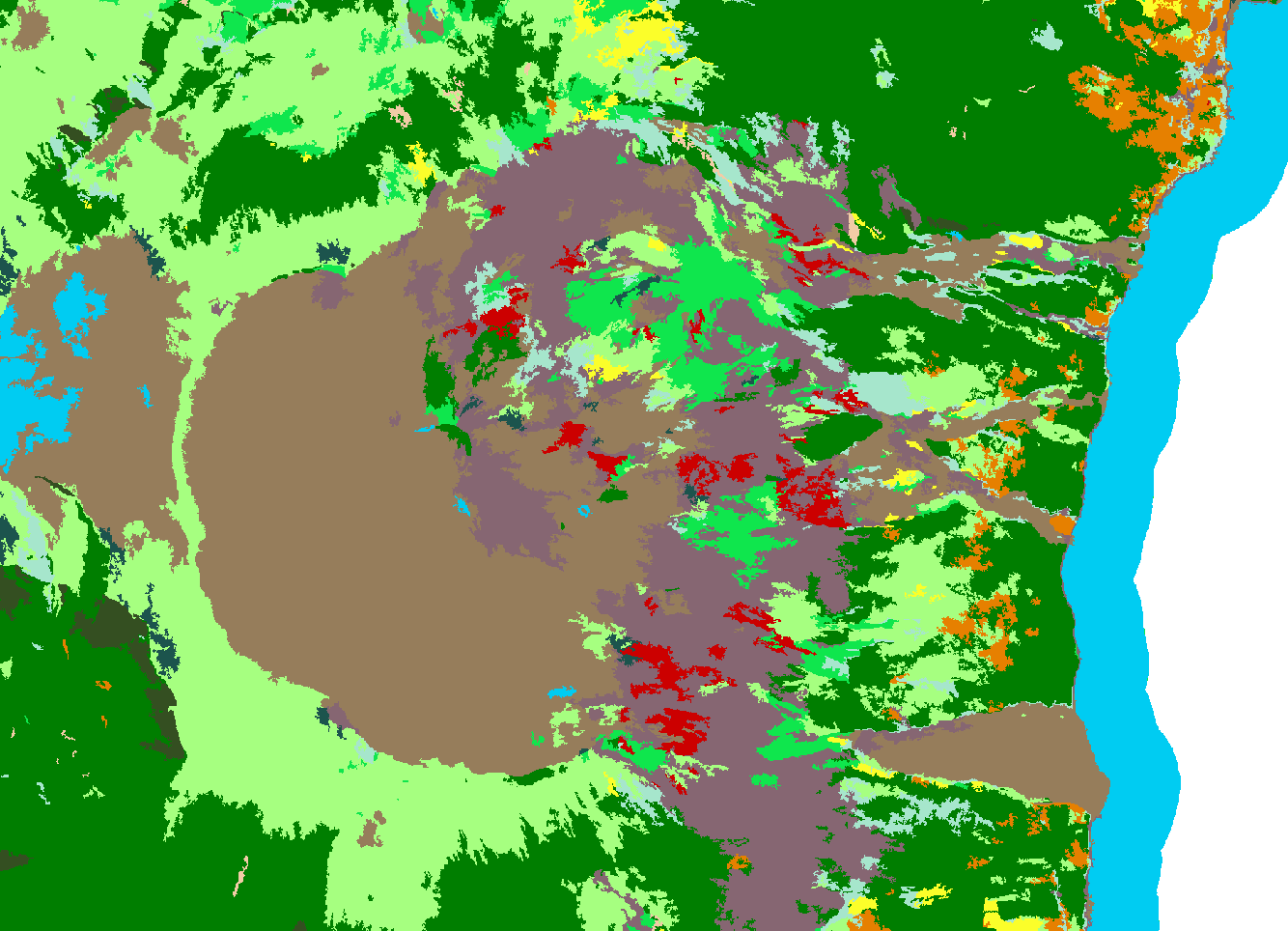} } & \subfloat[\label{fig:reunion_ex1_b}]{\includegraphics[width=.2\linewidth]{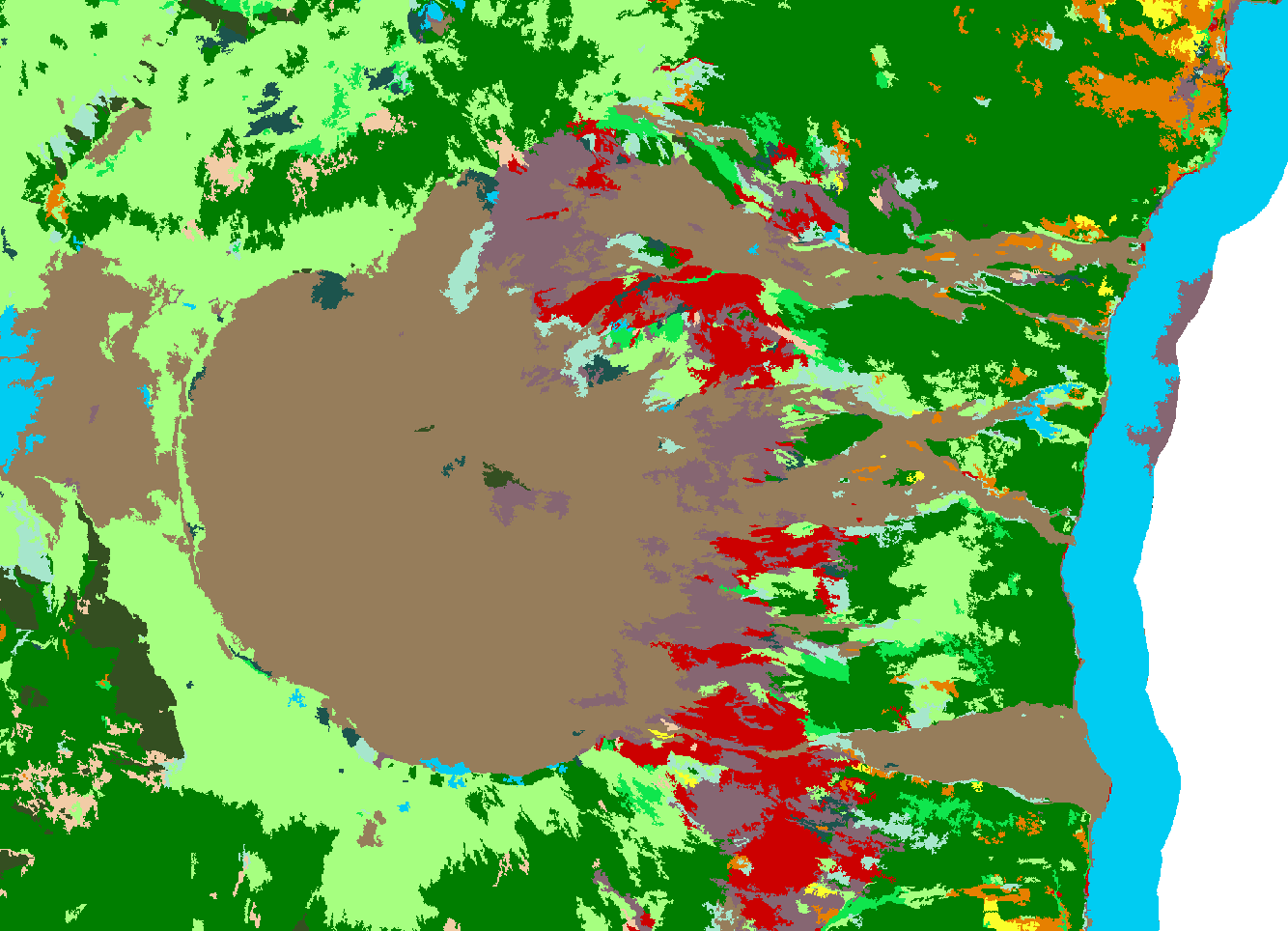} } \\
\multicolumn{4}{c}{ \includegraphics[width=.9\linewidth]{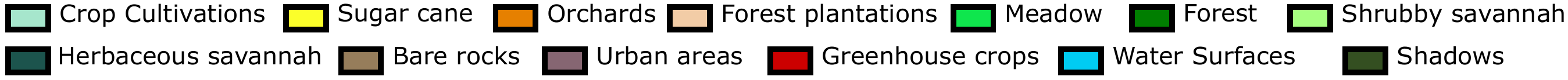}}

\end{tabular}
\caption{Qualitative investigation of Land Cover Map details produced by \textit{RF(S2)}, \textit{RF(S1,S2)} and \method{}  on two different zones (from the top to the bottom): i) an agricultural and ii) the volcano area.~\label{tab:reunion_examples}.}
\end{figure*}

Figure~\ref{tab:reunion_examples} reports two representative map classification details corresponding to the classification produced by the $RF(S2)$, $RF(S1,S2)$ and \method{} methods.  
To supply a reference image with natural colors, we here use the SPOT6/7 image used for object layer extraction. The first detail (top of the figure) focuses on a zone mainly characterized by \textit{Orchards} cultivation. Here, we can note that $RF(S2)$ tends to confuse $0$--\textit{Crop Cultivation} and $2$--\textit{Orchards} classes, with the latter often overestimated to the detriment of the former. Such classes are hardly distinguishable on optical imagery at 10~m resolution. This effect is reduced on the map provided by \method{}, probably due to the fact that it effectively exploits the information on the canopy structure provided by radar imagery. The second detail (bottom of the figure) depicts the \textit{Piton de la Fournaise} volcano on the eastern side of the island. The area between the volcano and the sea is often affected by the presence of clouds at S2 acquisition time. Due to this phenomenon, Random Forest models generate vast erroneous areas, and even $RF(S1,S2)$ is not able to balance out the lack of S2 imagery by exploiting S1 information. Conversely, the error is significantly reduced when using \method{}, demonstrating its ability to intelligently combine radar and optical time series, thus resulting in a better visual classification. Complete land cover maps are available online~\footnote{ \url{https://193.48.189.134/index.php/view/map/?repository=od2rnn\&project=OD2RNN}}.

\bibliographystyle{IEEEBib}
\footnotesize
\bibliography{biblio,dualHead}

\end{document}